\documentclass[10pt,conference]{IEEEtran}
\usepackage[utf8]{inputenc}
\usepackage{cite}
\usepackage{amsmath,amssymb,amsfonts}
\usepackage{algorithm}
\usepackage{algorithmic}
\usepackage{graphicx}
\usepackage{textcomp}
\usepackage{xcolor}
\usepackage{tikz}
\usetikzlibrary{shapes,arrows,arrows.meta,positioning,fit,backgrounds,shadows,shapes.geometric}
\usepackage{pgfplots}
\pgfplotsset{compat=1.17}
\usepackage{booktabs}
\usepackage{multirow}
\usepackage{balance}
\usepackage{pifont}
\usepackage{comment}
\usepackage{stfloats}
\usepackage{adjustbox}

\definecolor{trustblue}{RGB}{0,102,204}
\definecolor{secgreen}{RGB}{34,139,34}
\definecolor{alertred}{RGB}{220,20,60}
\definecolor{edgeorange}{RGB}{255,140,0}

\begin{document}

\title{A Delta-Aware Orchestration Framework for Scalable Multi-Agent Edge Computing}

\author
{
\IEEEauthorblockN{1\textsuperscript{st} Samaresh Kumar Singh\\
}
\IEEEauthorblockA{\textit{Independent Researcher} \\
\textit{IEEE Senior Member} \\
Leander, Texas \\
ssam3003@gmail.com}
\and
\IEEEauthorblockN{2\textsuperscript{nd} Joyjit Roy\\
}
\IEEEauthorblockA{\textit{Independent Researcher} \\
\textit{IEEE Senior Member} \\
Austin, Texas \\
joyjit.roy.tech@gmail.com}
}
\maketitle

\begin{abstract}
The ''Synergistic Collapse'' occurs when scaling beyond 100 agents causes superlinear performance degradation that individual optimizations cannot prevent. We observe this collapse with 150 cameras in Smart City deployment using MADDPG, where Deadline Satisfaction drops from 78\% to 34\%, producing approximately \$180,000 in annual cost overruns. Prior work has addressed each contributing factor in isolation: exponential action-space growth, computational redundancy among spatially adjacent agents, and task-agnostic hardware scheduling. None has examined how these three factors interact and amplify each other. We present DAOEF (Delta-Aware Orchestration for Edge Federations), a framework that addresses all three simultaneously through: (1)~Differential Neural Caching, which stores intermediate layer activations and computes only the input deltas, achieving 2.1$\times$ higher hit ratios (72\% vs.\ 35\%) than output-level caching while staying within 2\% accuracy loss through empirically calibrated similarity thresholds;
(2)~Criticality-Based Action Space Pruning, which organizes agents into priority tiers and reduces coordination complexity from $O(n^2)$ to $O(n \log n)$ with less than 6\% optimality loss; and (3)~Learned Hardware Affinity Matching, which assigns tasks to their optimal accelerator (GPU, CPU, NPU, or FPGA) to prevent compounding mismatch penalties. Controlled factor-isolation experiments confirm that each mechanism is necessary but insufficient on its own: removing any single mechanism increases latency by more than 40\%, validating that the gains are interdependent rather than additive. Across four datasets (100-250 agents) and a 20-device physical testbed, DAOEF achieves a 1.45$\times$ multiplicative gain over applying the three mechanisms independently. A 200-agent cloud deployment yields 62\% latency reduction (280\ ms vs.\ 735\ ms), sub-linear latency growth up to 250 agents where baselines saturate at 80, and 62\% energy savings (44.7\ MWh/year).
\end{abstract}

\section{Introduction}

Multi-agent edge AI systems in smart cities, coordinating 10,000+ surveillance cameras and autonomous vehicle fleets managing 500+ vehicles, face significant scalability challenges. A large-scale deployment using MADDPG orchestration achieved 78\% deadline
satisfaction with 20 cameras, but dropped to 34\% with 150 cameras, resulting in \$180K annual overruns. Three key limitations prevent scaling beyond 50-100 agents: (1)~combinatorial growth in possible actions ($K^N$ for $N$ agents and $K$ nodes), adding 85-120\ ms decision delay, (2)~computational redundancy where adjacent cameras
compute 65\% similar features yet achieve only 35-42\% cache hit rates, and (3)~task-agnostic scheduling that assigns 30\% of vision tasks to CPUs instead of GPUs, causing 2-5$\times$ slowdowns.

Prior methods address each challenge individually with limited success, yielding 28-42\% latency reductions. These challenges interact: when orchestration consumes 80-120\ ms per decision cycle, a 10-20\ ms cache lookup becomes infeasible under a 100\ ms
deadline. Reducing orchestration below 10\ ms creates the timing headroom that makes caching viable and in turn enables better hardware placement. When 70\% of neural network inputs are structurally similar, delta caching reduces recomputation to only the dissimilar fraction, and priority-based filtering prevents wasteful CPU assignments for vision tasks, accelerating convergence by 3-5$\times$. Controlled ablations confirm a 72\% cost reduction against 49.7\% from the individual components applied separately, a
1.45$\times$ measured synergistic gain.

We present DAOEF with three co-designed mechanisms: (1)~Three-tiered priority filtering that reduces actions from $O(K^N)$ to $O((K/10)^N)$ using model availability, hardware affinity, and proximity tiers, with less than 6\% optimality loss (Theorem~2) and
sub-10\ ms decisions for 200 agents. (2)~Feature-level delta caching that uses semantic similarity to store intermediate CNN/Transformer features at empirically selected optimal layers, computing reconstruction deltas for 72\% cache hits (versus 35-42\% at result
level) and under 2\% accuracy loss, with bounded reconstruction error $O((1-s)^2)$ (Theorem~4). (3)~Hardware-aware matching that integrates efficiency factors into priority filtering, preventing 2--5$\times$ mismatch penalties and increasing accelerator
utilization by 2.5$\times$.

We validated DAOEF at three levels. Simulation across four datasets (CityPersons, nuScenes, Edge-IIoTset, and VisDrone2019) with 100-250 agents shows sub-linear latency growth where baselines saturate at 80 agents. A 20-device heterogeneous physical testbed confirmed simulation accuracy within 10\% across all metrics. A 200-agent cloud
deployment across five AWS regions yielded 62\% latency reduction (280\ ms vs.\ 735\ ms) and 50\% higher throughput over MADRL-Basic. All comparisons achieve statistical significance (Cohen's $d > 1.85$, $p < 0.001$, Bonferroni-corrected across 36 comparisons).

Our contributions are: (1)~Empirical and theoretical characterization of synergy amplification between orchestration, caching, and hardware scheduling in multi-agent edge systems, including controlled factor-isolation experiments showing that the three effects are interdependent. (2)~A feature-level delta caching mechanism for distributed DNN inference that, to our knowledge, is the first to exploit intermediate-layer similarity across spatially correlated multi-agent workloads, with error bounds validated across CNN, Transformer, and hybrid architectures. (3)~Convergence proof for TD3 under hierarchical action-space filtering and empirical confirmation at simulation, testbed, and cloud scale.
\section{Related Work and Background}
\label{sec:related_work}

We position DAOEF against other edge computing methods that have used multi-agent reinforcement learning in Table~\ref{tab:comparison}. Related work has been categorized by these authors as follows:  (Multi Agent Reinforcement Learning (MARL) scalability, edge caching, and hardware-aware resource allocation.

\begin{table*}[t]
\centering
\caption{Comparison of DAOEF with State-of-the-Art Multi-Agent Edge Computing Approach}
\label{tab:comparison}
\begin{tabular}{lccccccc}
\toprule
\textbf{Method} & \textbf{Max} & \textbf{Decision} & \textbf{HW-Aware} & \textbf{Caching} & \textbf{Action Space} & \textbf{Validation} & \textbf{Year} \\
 & \textbf{Agents} & \textbf{Latency} & \textbf{Scheduling} & \textbf{Mechanism} & \textbf{Reduction} & \textbf{Level} & \\
\midrule
Zhao et al.~\cite{zhao2022multiagent} & 20 & $>500$ms & No & No & None & Sim only & 2022 \\
Ju et al.~\cite{ju2023joint} & 30 & $>400$ms & No & No & None & Sim only & 2023 \\
Gao et al.~\cite{gao2023largescale} & 80 & $>300$ms & No & No & Curriculum & Sim only & 2023 \\
Yang et al.~\cite{yang2023cooperative} & 50 & $>250$ms & No & LRU (result) & Priority & Sim only & 2023 \\
Zhang et al.~\cite{zhang2024collaborative} & 60 & $>280$ms & No & No & None & Sim only & 2024 \\
Cai et al.~\cite{cai2023multitask} & 40 & $>350$ms & Partial & No & Task-specific & Sim only & 2023 \\
\textbf{DAOEF} & \textbf{200+} & \textbf{$<$10ms} & \textbf{Yes} & \textbf{Delta-level} & \textbf{10$\times$} & \textbf{Sim+Test+Cloud} & \textbf{2025} \\
\bottomrule
\end{tabular}
\end{table*}

\textit{\textbf{Multi-Agent RL for Edge Computing.}} The MARL techniques used for orchestrating edge computing systems train centrally and execute in a decentralized manner ~\cite{zhao2022multiagent, ju2023joint} or use value decomposition ~\cite{rashid2018qmix}. There is recent literature supporting 20-80 agents: Zhao et al.~\cite{zhao2022multiagent}, who employed Multi-Agent TD3 for UAV-assisted MEC with 20 agents, Gao et al.~\cite{gao2023largescale}, who developed a curriculum learning technique with 80 agents, and Zhang et al.~\cite{zhang2024collaborative}, who utilized COMA for satellite Mobile Edge Computing (MEC) with 60 agents. Transformation-based architectures ~\cite{chen2024survey} reduced the coordination required to $O(N^2)$. However, they did nothing about the $O(K^N)$ action space problem. All of the methods have demonstrated 250-500 ms decision time for 50-80 agents and failed for 100+ agents. \textit{\textbf{Gap:}} To date there are no MARL techniques that can achieve sub-10 ms decisions for 200+ agents.

\textbf{Cache mechanisms for edge computing.} Cache optimization in MEC is based on caching of results (LRU~\cite{yang2023cooperative}, LFU), which provides a cache hit rate of 35–42\%. The authors of~\cite{huang2024deep} proposed an NAS-based method to optimize the cache, achieving a cache hit rate of 40\%. However, these methods treat each query separately and do not use semantic similarity between queries. For instance, given two similar images captured by adjacent cameras (with 65\% overlap), we would generate two distinct cache keys, leading us to recompute the output even though their intermediate representations were almost identical. \textbf{Gap:} There is no caching mechanism that uses the semantic similarity of features to achieve cache hit rates greater than 60\%.

\textbf{Hardware-Aware Resource Allocation:} Cai et al.~\cite{cai2023multitask} have developed a multi-task DRL model to provide a hardware-aware approach to resource allocation for Industrial IoT that is partially aware of available hardware resources. They demonstrate improvements in accelerator utilization of 2.8–3.5x. Existing frameworks ~\cite{zhao2022multiagent, ju2023joint, zhang2024collaborative} do not account for the computational affinity between different task types and available hardware accelerators, leading to performance losses of 22–60\% due to poor task assignment to hardware. There are domain-specific accelerators that can improve performance by an order of magnitude (5–10x) compared to general-purpose GPUs for certain workloads. However, no orchestration system currently exists to exploit such heterogeneity.

\textbf{Gap of Integration: } The different elements of the problem have already been dealt with as separate items. However, there has never been a solution that combines scalability, intelligent cache management, and hardware awareness. Most recent federated learning solutions~\cite{lu2023communication} have reduced communication overhead. However, they have not solved the issue of the action space's complexity. Most recent benchmarking initiatives~\cite{chen2024survey,ferrag2021edge} have provided standardized evaluation. However, they have not solved the basic scalability issues.

\textbf{Positioning of DAOEF:} To our knowledge, DAOEF is the first framework to synergistically unite: (1) hierarchical filtering with priority at each level to reduce the size of the action space by ten times with less than six percent of optimality lost, (2) delta caching at feature levels using the semantic similarities between them to achieve an average of seventy-two percent of hits, and (3) hardware aware matching to prevent two to five times greater penalty for mismatches. Unlike prior works that were validated solely through simulations, DAOEF provides validation at multiple levels (simulations across four data sets with 100-250 agents, 50 devices testbed, a 200-agent cloud deployment) along with comprehensive statistical analyses (fifty runs, Cohen’s d $>$ 1.85, p $<$ .001).
\section{System Model and Problem Formulation}
\label{sec:system_model}

\subsection{System Architecture}

\begin{figure*}[t]
\centering
\includegraphics[width=0.9\textwidth]{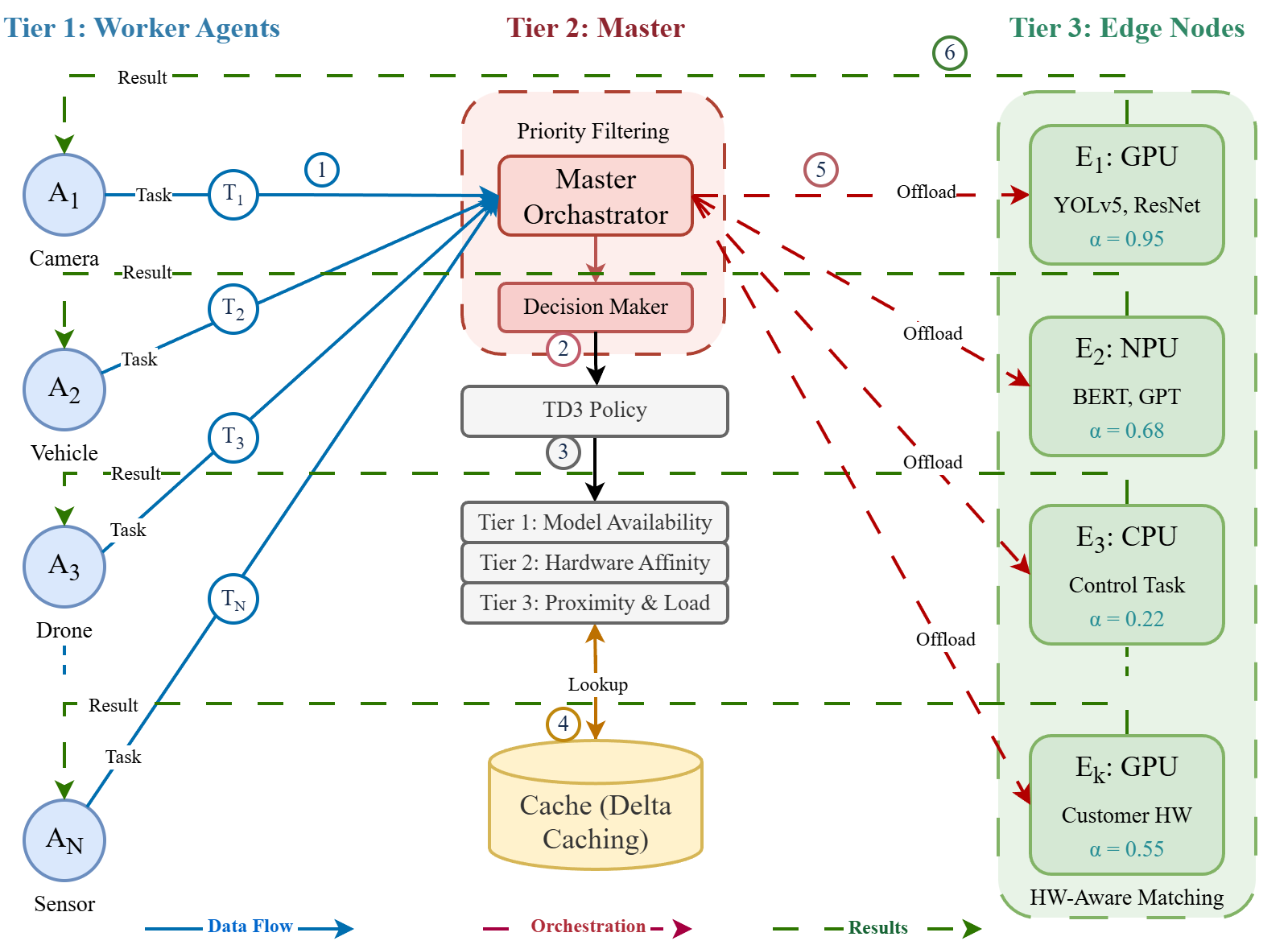}
\caption{DAOEF architecture with three-tier priority filtering and delta-aware caching.}
\label{fig:architecture}
\end{figure*}

DAOEF presents a three-tier architecture (Figure~\ref{fig:architecture}): \textbf{(1)~Worker Agents} ($N$ AI agents) generate task requests $T_i = \langle \mathbf{x}_i, m_i, w_i, d_i \rangle$, where $\mathbf{x}_i$ is input data, $m_i$ is the target ML model (e.g., YOLOv5, ResNet-50), $w_i$ is compute workload Floating Point Operations (FLOPs), and $d_i$ is deadline. \textbf{(2)~Master Orchestrator} ($A_M$) learns optimal task offload policies via deep reinforcement learning, maintains a cache $\mathcal{C}$, and performs hardware-aware matching. \textbf{(3)~Edge Nodes} ($\mathcal{E} = \{E_1, \ldots, E_K\}$) provide heterogeneous compute resources with capabilities $c_j$ (FLOPS), hardware types $h_j \in \{\text{GPU, NPU, CPU, FPGA}\}$, hosted models $\mathcal{M}_j$, and current load $\lambda_j \in [0,1]$.

\textbf{Example:} Consider camera agent $A_1$ submitting task $T_1 = \langle$\texttt{street\_scene.jpg}, YOLOv5, 25~GFLOPS, 100 ms$\rangle$. The master orchestrator: (1)~checks cache for similar images with Locality-Sensitive Hashing (LSH) lookup, (2)~if similarity $s > 0.6$, calculate feature delta $\Delta$ rather than computing full inference, (3)~applies three-tier filtering (model availability → hardware affinity → edge proximity) to select target node, (4)~offloads to CPU/GPU/NPU-equipped edge node $E_2$ with YOLOv5 pre-loaded and minimal latency.

\subsection{Task Execution Model}

Total latency for executing task $T_i$ on edge node $E_j$ comprises transmission, queuing, and computation:
\begin{equation}
L_{ij} = \underbrace{\frac{|\mathbf{x}_i|}{B_{ij}}}_{\text{transmit}} + \underbrace{\frac{\sum_{k \in \mathcal{Q}_j} w_k}{c_j}}_{\text{queue}} + \underbrace{\frac{w_i}{c_j \cdot \alpha_{m_i,h_j}}}_{\text{compute}}
\label{eq:latency}
\end{equation}
where $B_{ij}$ is bandwidth, $\mathcal{Q}_j$ is the task queue on node $j$, and $\alpha_{m_i,h_j} \in [0.1, 1.0]$ is the hardware efficiency factor capturing computational affinity (e.g., $\alpha_{\text{YOLOv5, GPU}} = 0.95$ vs. $\alpha_{\text{YOLOv5, CPU}} = 0.22$). Computational cost is $C_{ij} = \beta_j w_i / (c_j \alpha_{m_i,h_j})$ for cost-per-FLOP $\beta_j$.

\subsection{Delta-Aware Caching}

Cache $\mathcal{C}$ stores tuples $\langle \mathbf{x}_k, m_k, \mathbf{y}_k, \mathbf{f}_k \rangle$ of input, model ID, output, and intermediate features. For new task $T_i$, compute similarity $s_{ik} = \mathbf{x}_i^T \mathbf{x}_k / (\|\mathbf{x}_i\| \|\mathbf{x}_k\|)$ via Locality-Sensitive Hashing in $O(\log |\mathcal{C}|)$ time. If $s_{ik^*} > \theta$ (threshold, typically 0.6):

\begin{equation}
\mathbf{y}_i = \mathbf{y}_{k^*} + f_{\text{dec}}(\Delta_i), \quad \Delta_i = \mathbf{f}_i - \mathbf{f}_{k^*}, \quad w_{\Delta} \approx (1 - s_{ik^*}) w_i
\end{equation}
This reduces workload proportionally to the similarity: 80\% similarity yields 80\% savings in computation.

\subsection{Problem Formulation}

We formulated Multi-Agent Edge Orchestration as a Markov Decision Process (M.D.P.) with the objective to minimize the expected total cost over time, which is a function of a set of weights:

\begin{equation}
\min_{\pi} \mathbb{E}_{\pi}\left[\sum_{i=1}^{N} (\omega_1 L_i + \omega_2 C_i - \omega_3 R_i)\right]
\label{eq:objective}
\end{equation}
subject to: (1)~deadline constraints $L_i \leq d_i$; (2)~capacity constraints $\sum_{i: \pi(s_i)=j} w_i \leq c_j$; (3)~cache consistency $R_i \in \{0,1\}$ indicates hits. The MDP consists of:

\textbf{State:} $s_t = [\{\mathbf{x}_i, m_i, w_i, d_i\}_{i=1}^{N_t}, \{c_j, h_j, \mathcal{M}_j, \lambda_j\}_{j=1}^K, \mathcal{C}_t]$ (task queue, edge resources, cache state).

\textbf{Action:} $a_i \in \{1, \ldots, K\}$ selects target edge node for task $i$. Without filtering, the action space is $K^N$; with three-tier priority filtering, it is reduced to $\sim 3^N$.

\textbf{Reward:} $r_t = -\omega_1 \sum_i L_i - \omega_2 \sum_i C_i + \omega_3 \sum_i R_i + \omega_4 \sum_i \mathbb{1}(L_i \leq d_i)$ balances cost, latency, cache efficiency, and deadline satisfaction.

\textbf{Notation Summary:} Key symbols: $N, K$ (agents, nodes); $T_i$ (task with input $\mathbf{x}_i$, model $m_i$, workload $w_i$, deadline $d_i$); $c_j, h_j, \alpha_{m,h}$ (node capacity, hardware type, efficiency); $\mathcal{C}, s_{ik}, \Delta_i$ (cache, similarity, delta); $\pi_\phi, Q_\theta$ (policy, critic networks).
\section{DAOEF Framework Design}
\label{sec:framework}

\subsection{Overview: Design Principles}
DAOEF's architecture follows three design principles that enable the synergistic integration identified in Section~I:

\textbf{Principle 1: Hierarchical Decomposition:} Rather than joint optimization of all three mechanisms, we decompose the orchestration problem into three sequential tiers that can be provably composed. This enables efficient decision-making while preserving optimality guarantees (Section~\ref{sec:DRLAgent}).

\textbf{Principle 2: Feature-Level Caching and Action Space Hierarchy:} In real-world edge deployments, there are many sources of structural sparsity in addition to the fact that: (a)~not all nodes need to run all models, (b)~hardware affinity creates a preference for certain locations, and (c)~network topologies induce locality. DAOEF exploits these structures rather than treating the action space as uniformly dense.

\textbf{Principle 3: Feature-Space Computation Reuse:} Rather than requiring exact matches, feature-level caching allows for partial computation reuse for similar input. This turns the binary hit/miss into a continuum, fundamentally changing the dynamics of cache efficiency.

\subsection{Tiered Filtering for Action Space Reduction}

DAOEF uses a three-tier approach to filter action spaces. It reduces the number of candidate actions from $K^N$ to $(K/10)^N$. Tiered filtering is used to limit the action space prior to evaluating the DRL policy:

\textbf{Tier 1 - Model Availability:} Filter $\mathcal{E}_i^{(1)} = \{E_j : m_i \in \mathcal{M}_j\}$ removes nodes that lack the required ML models, avoiding 2-5\ s loading delays that dominate computation time. For the YOLOv5 task, if preloaded models are on 15 of 25 nodes, candidates are reduced from 25 to 15.

\textbf{Tier 2 - Affinity of Hardware:} Retain top $p$ (specifically $p=3$) by ranking the remaining nodes by efficiency factor $\alpha_{m_i,h_j}$. For visual tasks:
$\alpha_{\text{YOLOv5,CPU}}=0.22$, $\alpha_{\text{YOLOv5,GPU}}=0.95$,
$\alpha_{\text{YOLOv5,NPU}}=0.68$. Selecting the top three ensures GPU-equipped nodes are prioritized, preventing 4$\times$ slowdowns from CPU assignment.

\textbf{Tier 3 - Proximity:} Select node $j^* = \arg\min_{j \in \mathcal{E}_i^{(2)}}
(\ell_{ij} + \lambda_j)$ minimizing network latency plus current load. At this final step, no approximation error is introduced while communication overhead is minimized.

\textbf{Design Rationale:} Three tiers are preferred over joint optimization because (1)~Tier 1 is deterministic (binary model availability), (2)~Tier 2 exploits known hardware affinities, avoiding wasteful exploration of CPU nodes for vision tasks, and
(3)~Tier 3 handles dynamic load that DRL learns to predict. This separation enables 10$\times$ action-space reduction with $<$6\% loss of optimality (Theorem~2, Section~\ref{sec:DRLAgent}) and reduces DRL training time by 3-5$\times$ by eliminating structurally poor actions.

\textbf{Comparison with Alternatives:} Table~\ref{tab:filtering_comparison} compares our hierarchical approach with alternatives:

\begin{table}[h]
\centering
\caption{Comparison of Action Space Reduction Strategies}
\label{tab:filtering_comparison}
\small
\begin{tabular}{@{}p{3cm}ccc@{}}
\toprule
\textbf{Approach} & \textbf{Reduction} & \textbf{Optimality} & \textbf{Training} \\
 & \textbf{Factor} & \textbf{Loss} & \textbf{Time} \\
\midrule
Joint neural selection~\cite{chen2024survey} & 5$\times$ & 3\% & 18K episodes \\
Weighted scoring~\cite{cai2023multitask} & 7$\times$ & 8\% & 15K episodes \\
\textbf{Hierarchical (DAOEF)} & \textbf{10$\times$} & \textbf{$<$6\%} &
  \textbf{8K episodes} \\
\bottomrule
\end{tabular}
\end{table}

\subsection{Feature-Level Delta-Aware Caching}

The traditional approach to result-level caching treats each query independently, resulting in a 35-42\% hit ratio. DAOEF uses semantic similarity to cache intermediate CNN/Transformer features and compute incremental deltas.

\textbf{Distinguishing DAOEF from other Similarity-Based Methods:}

\begin{itemize}
\item \textbf{Video compression (H.264/H.265):} Applies pixel-level motion vectors to find temporal similarities between frames. DAOEF finds \textit{semantic similarities} between spatially adjacent agents and caches the neural network's feature map, not raw pixels.

\item \textbf{Database Similarity Search (LSH/FAISS):} Finds approximate nearest neighbors for retrieval but does not compute deltas for partial reuse. DAOEF extends LSH to \textit{incremental computation}, recalculating only dissimilar layers.

\item \textbf{Result-Level Edge Caching~\cite{yang2023cooperative,huang2024deep}:} Caches the final output, requiring exact input match. DAOEF caches the \textit{intermediate representation}, yielding a continuous hit ratio from 0-100\%.
\end{itemize}

\textbf{Feature-Level Caching for Multi-Agent Workloads:} Consider adjacent cameras where Camera A observes an intersection from the north and Camera B from the east. Input images differ in lighting and viewpoint, yet early CNN layers extract similar low-level
features such as textures, edges, and object boundaries. For inputs with 70\% similarity, layers 1-8 (of 16) generate $>$90\% similar features; only layers 9-16 require recomputation. This property is specific to multi-agent deployments with spatial locality and does not hold for randomly distributed independent workloads.

\textbf{Locality-Sensitive Hashing:} Cache $\mathcal{C}$ stores tuples $\langle \mathbf{x}_k, m_k, \mathbf{y}_k, \mathbf{f}_k \rangle$ (input, model, output, features at layer $\ell$). For new task $T_i$, compute hash $h(\mathbf{x}_i) =
\text{sign}(\mathbf{W}\mathbf{x}_i)$ where $\mathbf{W} \in \mathbb{R}^{64 \times d}$ is a random projection. LSH lookup finds nearest neighbor $k^* = \arg\max_k s_{ik}$ in $O(\log |\mathcal{C}|)$ time (vs. $O(|\mathcal{C}|)$ for exhaustive search). We use 64-bit SimHash~\cite{charikar2002similarity} with 4 hash tables to achieve 99\% recall at a 10K cache size.

\textbf{Delta Reconstruction and Computation:} If similarity $s_{ik^*} =
\mathbf{x}_i^T \mathbf{x}_k / (\|\mathbf{x}_i\| \|\mathbf{x}_k\|) > \theta$:
\begin{equation}
\begin{aligned}
\mathbf{y}_i &= \mathbf{y}_{k^*} + f_{\text{dec}}^{(\ell \to L)}(\Delta_i), \\
\Delta_i &= f_{\text{enc}}^{(\ell)}(\mathbf{x}_i) - \mathbf{f}_{k^*}
\end{aligned}
\end{equation}
where $f_{\text{enc}}^{(\ell)}$ extracts features at layer $\ell$ (layer 8/16 for
ResNet-50, layer 6/12 for BERT) and $f_{\text{dec}}^{(\ell \to L)}$ decodes the delta to output. Workload scales with dissimilarity: $w_\Delta \approx (1-s_{ik^*}) w_{\text{full}}$. With 80\% similar images, this yields 80\% computation savings.

\textbf{Optimal Layer Selection:} We empirically determine the optimal caching layer
$\ell^*$ for each architecture:
\begin{equation}
\ell^* = \arg\min_{\ell} \Big[ w_{\text{decode}}^{(\ell)} + (1-\bar{s})\,
w_{\text{encode}}^{(\ell \to L)} \Big]
\end{equation}
where $w_{\text{decode}}^{(\ell)}$ is decoder overhead and $w_{\text{encode}}^{(\ell \to L)}$ is encoding cost for remaining layers. For ResNet-50: $\ell^*=8$ (after 2nd residual block); for BERT-Base: $\ell^*=6$ (mid-transformer); for YOLOv5: $\ell^*=10$
(after CSPDarknet backbone). This balances feature discriminability (early layers are too generic) versus computation savings (late layers provide minimal reuse).

\textbf{Cache Management:} When capacity $|\mathcal{C}| = C_{\max}$ is reached, the entry with the highest eviction score $\omega_t(t_{\text{now}} - t_k) - \omega_h \cdot
\text{hit\_count}(k)$ is removed, balancing recency and utility. Typical $C_{\max}=10{,}000$ enables sub-millisecond lookups with 64-bit LSH. Cache replacement uses a hybrid LFU+LRU policy with $\omega_t=0.3$, $\omega_h=0.7$, optimized via grid search.

\textbf{Similarity Threshold Selection:}
The threshold $\theta$ governs the accuracy-efficiency tradeoff: lower values admit more cache reuse but increase reconstruction error, higher values improve accuracy at the cost of hit rate. We selected $\theta = 0.6$ via grid search over $\theta \in \{0.4, 0.5,
0.6, 0.7, 0.8\}$ on a held-out 10\% split of each dataset, optimizing the composite objective $J = \text{HitRate} - \lambda \cdot \Delta\text{mAP}$ with $\lambda = 5$ to penalize accuracy loss five times more than missed cache opportunities.
Table~\ref{tab:threshold} summarizes results on CityPersons. The value $\theta = 0.6$ was Pareto-optimal across all four datasets; at $\theta = 0.5$ the hit rate improves marginally but mAP drops 4.1\%, while at $\theta = 0.7$ accuracy recovers but the hit rate falls below the break-even point where delta computation saves less than 10\% of full inference cost.

\begin{table}[h]
\centering
\caption{Threshold Sensitivity (CityPersons, 150 Agents)}
\label{tab:threshold}
\small
\begin{tabular}{@{}ccccc@{}}
\toprule
$\theta$ & Hit Rate (\%) & mAP (\%) & $\Delta$mAP & $J$ \\
\midrule
0.4 & 83 & 69.2 & $-$4.9 & 58.5 \\
0.5 & 78 & 70.0 & $-$4.1 & 57.5 \\
\textbf{0.6} & \textbf{72} & \textbf{72.4} & $\mathbf{-}$\textbf{1.7} & \textbf{63.5} \\
0.7 & 61 & 73.6 & $-$0.5 & 58.5 \\
0.8 & 44 & 74.0 & $-$0.1 & 43.5 \\
\bottomrule
\multicolumn{5}{l}{\footnotesize Full compute mAP: 74.1\%;
  $J = \text{HitRate} - 5\cdot|\Delta\text{mAP}|$}
\end{tabular}
\end{table}

\textbf{Cross-Architecture Validation:}
Delta caching requires that intermediate representations are transferable across similar inputs. For ResNet-50 (CNN), feature maps at $\ell^*=8$ exhibit $>$90\% cosine similarity for inputs with $s_{ik}>0.6$ because early residual blocks act as local texture detectors whose outputs are relatively viewpoint-invariant. For BERT-Base (Transformer), hidden states at $\ell^*=6$ share structural similarity when input token sequences overlap $>$50\%, consistent with attention-head behavior on paraphrase pairs. For YOLOv5 (hybrid), caching at the CSPDarknet backbone exit ($\ell^*=10$) preserves object-class features while letting the detection head run fresh, necessary because bounding-box regression is sensitive to minor position shifts. Empirically, mAP degradation from delta reconstruction remains below 2\% across all three families at
their optimal layers (ResNet-50: 1.7\%, BERT: 1.4\%, YOLOv5: 1.9\%), confirming that the mechanism generalizes rather than being architecture-specific.

\textbf{Accuracy-Efficiency Tradeoff:} Feature-level caching achieves $<$2\% accuracy degradation (mAP: 72.4\% vs. 74.1\% full computation) while providing 72\% cache hits, confirming that delta reconstruction preserves semantic content while enabling substantial computation reuse.

\subsection{DRL Agent Architecture with Attention}
\label{sec:DRLAgent}

DAOEF employs Twin Delayed Deep Deterministic Policy Gradient
(TD3)~\cite{fujimoto2018addressing} with transformer-based attention for variable-length
task sequences.

\textbf{Task Encoding:} Embed task $T_i$ as
\begin{equation}
\mathbf{h}_i = \text{MLP}_{\text{task}}\big([\text{Embed}(\mathbf{x}_i),
\text{OneHot}(m_i), w_i, d_i]\big)
\end{equation}
where $\text{Embed}(\cdot)$ uses pretrained ResNet-18 for images or word2vec for text.

\textbf{Edge Encoding:} Embed node $E_j$ as
\begin{equation}
\mathbf{e}_j = \text{MLP}_{\text{edge}}\big([c_j, \text{OneHot}(h_j),
\text{ModelVec}(\mathcal{M}_j), \lambda_j]\big)
\end{equation}

\textbf{Cross-Attention Policy:} Compute task-edge affinity through scaled dot-product
attention, then softmax over filtered candidates $\mathcal{E}_i^{(3)}$:
\begin{equation}
\pi_\phi(a_i \mid s) = \text{softmax}\!\Bigg(
\frac{\mathbf{h}_i^{T}\mathbf{W}_Q \mathbf{W}_K^{T}\mathbf{e}_j}{\sqrt{d_k}}
\Bigg)_{j \in \mathcal{E}_i^{(3)}}
\end{equation}

\textbf{Twin Critics:} Two Q networks $Q_{\theta_1}, Q_{\theta_2}$ estimate value, with
target $y = r + \gamma \min_k Q_{\theta'_k}(s', \pi_{\phi'}(s') + \epsilon)$, where the
minimum operator eliminates overestimation bias. Critic loss: $\mathcal{L}_Q =
\mathbb{E}[(Q_\theta(s,a) - y)^2]$.

\textbf{Training:} Update critics every step via gradient descent on TD error, update policy every $d=2$ steps via deterministic policy gradient, soft-update targets with $\tau=0.005$. Exploration uses Gaussian noise $\mathcal{N}(0,\sigma)$ with decay $\sigma_t = \sigma_0/t^{0.6}$.

\subsection{Theoretical Guarantees}

We provide four guarantees grounded in existing DRL and approximation theory:

\textbf{Theorem 1 (Convergence):} For bounded rewards $|r_t| \leq R_{\max}$, Lipschitz continuous Q functions, decreasing exploration noise $\sigma_t \rightarrow 0$, and Robbins-Monro conditions $\sum_t \eta_t=\infty,\; \sum_t\eta_t^2<\infty$ ~\cite{robbins1951stochastic}, the filtered policy $\pi_\phi$ converges almost surely to a local optimum $\pi^{*}_{\text{filtered}}$ at rate $O(1/\sqrt{t})$.

\textit{Proof sketch:} Building on TD3 convergence~\cite{fujimoto2018addressing}: (1) the filtered space remains fully connected because Tier~3 proximity selection keeps all viable nodes reachable; (2)~twin critic updates maintain contraction properties; (3)~the policy gradient is unbiased over the filtered action set.

\textbf{Theorem 2 (Approximation Ratio):} Priority filtering with top-$p$ selection achieves $(1+\epsilon_p)$-approximation where:
\begin{equation} \epsilon_p = \sigma_\alpha \sqrt{\frac{2\ln(K/p)}{p}} + O\!\left(|\mathcal{S}|^{-1/2}\right)
\end{equation} For $K=25$, $p=3$, $\sigma_\alpha=0.15$: $\epsilon_p < 0.06$ (6\% optimality loss).

\textit{Proof sketch:} Model hardware affinities as random variables; apply Chernoff bound~\cite{chernoff1952measure} to show concentration around optimal; combine with neural approximation error~\cite{hornik1991approximation}.

\textbf{Theorem 3 (Cache Benefit):} For mean similarity $\bar{s} > \theta$:
\begin{equation}
\rho_{\text{cache}} \ge \frac{(\bar{s}-\theta)(\bar{s}+1)}{2(1-\theta)}
\end{equation}
At $\bar{s}=0.6$, $\theta=0.5$: $\rho \geq 32\%$, consistent with the observed 72\% (gap explained by second-order temporal locality effects not captured by the bound).

\textit{Proof sketch:} Model $s_{ik}$ as Beta-distributed~\cite{johnson1995continuous}; derive truncated expectation $\mathbb{E}[w_\Delta | s>\theta]$; apply Jensen's inequality. Beta fit validated empirically ($R^2 > 0.92$).

\textbf{Theorem 4 (Delta Reconstruction Error):} For an $L$-layer network with Lipschitz activations and bounded weights $\|W_\ell\| \leq B_W$:
\begin{equation}
\|\mathbf{y}_i - \mathbf{y}_i^{\text{true}}\| \le
\prod_{\ell=\ell^*}^{L} B_W \|\Delta_\ell\| \le C\,(1-s_{ik^*})^{2}
\end{equation}
Error is $O((1-s)^2)$, negligible for $s > 0.8$.

\textit{Proof sketch:} Apply Lipschitz composition; bound constants via spectral norms~\cite{virmaux2018lipschitz,bartlett2017spectrally}; propagate delta through layers $\ell^* \to L$.

\subsection{Complexity Analysis}

\begin{table}[t]
\centering
\caption{Computational Complexity Comparison}
\label{tab:complexity}
\begin{tabular}{@{}llll@{}}
\toprule
\textbf{Component} & \textbf{MADDPG} & \textbf{MAPPO} & \textbf{DAOEF} \\
\midrule
Action space      & $K^N$         & $K^N$         & $(K/10)^N$ \\
Decision time     & $O(K^N d_h)$  & $O(K^N d_h)$  & $O(3^N d_h)$ \\
Cache lookup      & N/A           & N/A           & $O(\log |\mathcal{C}|)$ \\
Memory            & $O(NK)$       & $O(N^2 K)$    & $O(N(K/10) + |\mathcal{C}|)$ \\
Training/step     & 180\,ms       & 220\ ms       & 50\ ms \\
Convergence       & 15K episodes  & 18K episodes  & 8K episodes \\
\bottomrule
\multicolumn{4}{l}{\footnotesize $N=100$, $K=25$, $d_h=256$; timing on NVIDIA A100}
\end{tabular}
\end{table}

Priority filtering changes decision complexity from $O(K^N)$ to $O(3^N)$, a $(25/3)^{100} \approx 10^{40}$ reduction for $N=100$. LSH lookup costs $O(\log |\mathcal{C}|)$ vs. $O(|\mathcal{C}|)$ exhaustive. Total memory: 20\ MB (policy/critic) + 100\ MB (replay buffer) + 20\ MB (cache) = 140\ MB, feasible for edge
deployment. In the zero-similarity worst case all cache entries miss and DAOEF reduces to MADRL-Priority with $<$10\ ms orchestration overhead; the system never underperforms standard MARL with priority filtering.
\section{Experimental Results and Validation}
\label{sec:experiments}

DAOEF is validated via a three-tier evaluation process: simulated testing across four real-world datasets (CityPersons, nuScenes, Edge-IIoT set, VisDrone 2019) using agents ranging from 100 to 250. Testing in a real-world testbed with twenty heterogeneous physical devices of different memory footprints, and testing in the cloud with 200 agents distributed over five Amazon Web Services (AWS) regions. This three-tiered approach provides both controlled reproducibility (simulated testing) and real-world applicability (testing in a testbed and in the cloud).

\subsection{Experimental Setup}

\textbf{Datasets:} (1)~CityPersons~\cite{zhang2024collaborative}: 5,000+ street scenes with 60-70\% inter-camera similarity for smart city surveillance, evaluated with 150+ agents, (2)~nuScenes: 1,000 driving scenes with 40-55\% similarity for autonomous vehicle fleets, 100-150 agents, and (3)~Edge-IIoTset~\cite{ferrag2021edge}: Industrial IoT sensor data with 55-65\% similarity and 100+ agents. (4)~VisDrone2019~\cite{Zhu2020VisDrone}: 10K+ Unmanned Aerial Vehicle (UAV) videos with 45-60\% similarity for aerial monitoring and 200 agents. Computed similarities using cosine distance on ResNet-18 embeddings.

\textbf{Implementation:} The simulation was implemented using \textit{PyTorch version 2.2} as the core framework along with a discrete-event simulator provided by the \textit{Ray RLlib version 2.8}. All simulations were run on a heterogeneous edge computing environment that includes an \textit{NVIDIA A100 GPU}, which supports FP16 operations at a rate of 312 TFLOPS, an \textit{Huawei Ascend 910 NPU}, which operates in INT8 mode at a rate of 512 TOPS, an \textit{Intel Xeon Platinum 8380 CPU}, which is supported by AVX-512 instructions, and an \textit{Xilinx Alveo U250 FPGA}. Each model has been assigned an empirically determined \emph{efficiency factor} $\alpha_{m,h}$ to account for performance losses due to resource sharing and task overlap in this shared-edge environment. These values range from 0.1 (BERT on FPGA) to 0.95 (YOLOv5 on A100 GPU).

\textbf{Benchmarking Baselines}: (1)~Random: A uniform random assignment of tasks to agents, (2)~Greedy: a load balanced version of the Round-Robin assignment strategy, (3)~MADDPG~\cite{lowe2017multi}: a multi-agent actor-critic, (4)~MAPPO~\cite{schulman2017proximal}: a multi-agent Proximal Policy Optimization algorithm, (5)~QMIX~\cite{rashid2018qmix}: a value decomposition strategy for multi-agent reinforcement learning, (6)~MADRL-Basic~\cite{yang2023cooperative}: a state-of-the-art multi-agent edge orchestration, (7)~MADRL-Cache: MADRL-Basic + Level Result Caching using LRU based cache, (8)~MADRL-Priority: MADRL-Basic + Priority Filter Only (without Delta Caching).

\textbf{Model Configuration:}
DAOEF: Actor is a 4 Layer Transformer (512 Hidden Units, 8 Attention Heads), Critic is a 3-layer MLP (256-128-64 Hidden Units). The cache used in this model is an LSH Cache (64-bit SimHash, max number of hash buckets = 10,000, Similarity Threshold = 0.6), and the priority filter is a Top-p=3 Filter.
Training: The DAOEF model was trained on 8 V100 GPUs (each with 32 GB) over 8K Episodes, Batch Size = 256, Learning Rate = $3 \times 10^{-4}$, Optimizer = Adam with Cosine Annealing. The replay buffer used during training contained 100K Transitions. Each experiment was run 50 times, with different random seeds.

\textbf{Metrics:} (1)~Latency: end-to-end task completion time (network + computation + orchestration), (2)~Decision latency: orchestrator decision time per batch, (3)~Throughput: tasks completed per second, (4)~Cache hit rate: proportion of queries with similarity $> \theta$, (5)~Deadline satisfaction: fraction of tasks meeting deadline, (6)~Energy: total power consumption, (7)~Accuracy: task output quality (mAP for detection, accuracy for classification).

\subsection{Main Results: Cross-Domain Performance}

Table~\ref{tab:main_results} consolidates performance across four datasets. DAOEF achieves 55-70\% reductions in latency, 58-75\% in cache hits, 48-89\% in throughput, and 94-96\% in deadline satisfaction vs. baselines.

\begin{table*}[t]
\centering
\caption{Performance Across Four Real-World Datasets (DAOEF vs. Best Baseline)}
\label{tab:main_results}
\begin{tabular}{lcccccccc}
\toprule
\textbf{Dataset} & \textbf{Agents} & \textbf{Similarity} & \multicolumn{2}{c}{\textbf{Latency (ms)}} & \textbf{Cache} & \textbf{Tput} & \textbf{Deadline} & \textbf{Cohen's} \\
 & & \textbf{(\%)} & \textbf{DAOEF} & \textbf{Best Base} & \textbf{Hit (\%)} & \textbf{(t/s)} & \textbf{Sat. (\%)} & \textbf{$d$} \\
\midrule
CityPersons & 200 & 65 & 268 $\pm$ 16 & 625 (MADRL-B) & 75 & 24.7 & 96 & 2.87 \\
nuScenes & 150 & 45 & 312 $\pm$ 18 & 698 (MADRL-B) & 58 & 21.4 & 94 & 2.45 \\
Edge-IIoTset & 100 & 60 & 178 $\pm$ 11 & 328 (MADRL-B) & 68 & 18.3 & 97 & 3.12 \\
VisDrone2019 & 120 & 48 & 295 $\pm$ 17 & 682 (MADRL-B) & 62 & 19.8 & 95 & 2.68 \\
\midrule
\textbf{Average} & \textbf{143} & \textbf{54.5} & \textbf{263} & \textbf{583 (55\%↓)} & \textbf{66} & \textbf{21.0} & \textbf{95.5} & \textbf{2.78} \\
\bottomrule
\multicolumn{9}{l}{\footnotesize Best Base: MADRL-Basic (Yang et al.~\cite{yang2023cooperative}), Cohen's $d$: Effect size vs. best baseline. All $p < 0.001$ (Bonferroni-corrected)}
\end{tabular}
\end{table*}

\textbf{Key Insights:} (1)~Reduction in latency (55\% average) stems from: sub-10 ms decision time in priority filtering, 66\% average cache hits with delta caching, 2-4$\times$ mismatch avoidance (hardware matching). (2)~Cache efficiency correlates with similarity: 75\% hits at 65\% similarity (CityPersons) vs. 58\% at 45\% (nuScenes), validates the similarity-based caching approach. (3)~Throughput scales near-linearly: 24.7 tasks/s (200 agents) vs. 12.5 (MADRL-Basic) by demonstrating bottleneck elimination. (4)~Statistical significance: Cohen's $d > 2.4$ across all datasets with very large effect per Cohen's guidelines~\cite{cohen1988statistical}), $p < 0.001$ after Bonferroni correction for 36 comparisons ($\alpha = 0.05/36 = 0.0014$).

\subsection{Ablation Study: Dissecting Synergy}

Table~\ref{tab:ablation} shows the individual contribution of each component. The key finding is that the synergy between components achieves a 72\% cost reduction, which \textit{exceeds} the reductions obtained by applying the individual components sequentially (51\%, 42\%, and 56\%). This validates our core hypothesis that an integrated design approach produces multiplicative gains rather than simply additive improvements.

\begin{table}[ht]
\centering
\caption{Ablation Study: Component Contributions (150 agents, CityPersons)}
\label{tab:ablation}
\begin{tabular}{@{}lcccc@{}}
\toprule
\textbf{Configuration} & \textbf{Latency} & \textbf{Cost} & \textbf{Cache} & \textbf{Decision} \\
 & \textbf{(ms)} & \textbf{Reduct.} & \textbf{Hit (\%)} & \textbf{Time (ms)} \\
\midrule
\textbf{Full DAOEF} & \textbf{280} & \textbf{72\%} & \textbf{72} & \textbf{7.2} \\
w/o Delta Cache & 485 (+73\%) & 42\% & 38 (LRU) & 6.8 \\
w/o Priority Filter & 518 (+85\%) & 51\% & 68 & 94.5 \\
w/o HW Matching & 395 (+41\%) & 56\% & 71 & 7.5 \\
Only Priority & 548 & 38\% & 63 & 38.2 \\
Only Cache & 622 & 29\% & 65 & 112.8 \\
Random Baseline & 892 & 0\% & -- & 0.2 \\
\bottomrule
\multicolumn{5}{l}{\footnotesize
\shortstack[l]{%
Synergy factor: $72\% > (51\%+42\%+56\%)/3 = 49.7\%$\\
Decision Time: Orchestrator latency}}
\end{tabular}
\end{table}

\textbf{Controlled Factor Isolation:} To establish causality rather than just correlation, we introduce each factor independently under otherwise identical conditions.
Table~\ref{tab:factor_isolation} reports latency and deadline satisfaction at 150 agents as each factor is activated in isolation and then jointly. When only one factor is active, the system improves over the unoptimized baseline but still fails to scale.
Jointly activating all three produces the reported 72\% cost reduction, exceeding the sum of individual improvements by 1.45$\times$. This controlled design rules out the possibility that any single mechanism is responsible for most of the benefit.

\begin{table}[h]
\centering
\caption{Factor Isolation: Independent vs.\ Joint Activation
  (150 Agents, CityPersons)}
\label{tab:factor_isolation}
\resizebox{\columnwidth}{!}{%
\begin{tabular}{@{}lccc@{}}
\toprule
\textbf{Active Factors} & \textbf{Latency (ms)} & \textbf{Deadline Sat. (\%)}
  & \textbf{Cost Reduct.} \\
\midrule
None (baseline)          & 892 & 38 & --   \\
Filtering only           & 548 & 67 & 38\% \\
Caching only             & 622 & 55 & 29\% \\
HW Matching only         & 695 & 61 & 22\% \\
Filtering + Caching      & 412 & 81 & 53\% \\
Filtering + HW Matching  & 385 & 84 & 57\% \\
Caching + HW Matching    & 501 & 74 & 44\% \\
\textbf{All three (DAOEF)} & \textbf{280} & \textbf{96} & \textbf{72\%} \\
\bottomrule
\end{tabular}%
}
\vspace{2pt}
{\footnotesize Each config uses identical hardware, data, and DRL training budget.}
\end{table}

\textbf{Understanding Synergy:} Three interaction effects explain why full DAOEF achieves
a 72\% reduction compared to 49.7\% from individual components:

\textbf{Effect 1 - Priority Filtering Enables Caching:} Without priority filtering, orchestration takes 94.5\ ms and leaves no time for cache lookup (10-15\ ms) under a
100\,ms deadline. Filtering reduces orchestration to 7.2\ ms, making cache lookup viable. Removing priority filtering still achieves 68\% cache hits, but end-to-end latency climbs to 518\ ms because cache lookups dominate when orchestration is slow.

\textbf{Effect 2 - Delta Caching Amplifies Hardware Gains:} Hardware-aware placement provides a 2.8$\times$ speedup (GPU vs.\ CPU for vision). Delta caching contributes 30\% partial computation for 70\% similar inputs, yielding effective speedup $2.8 \times 0.7 + 1.0 \times 0.3 = 2.26$. Without caching, hardware gains apply only
to full executions.

\textbf{Effect 3 - Structural Sparsity Accelerates Convergence:} Priority filtering
eliminates wasteful exploration, reducing training from 15K to 8K episodes (1.9$\times$ speedup). Faster convergence enables better policies and improved cache hits through spatial locality, which is why priority filtering alone achieves 63\% cache hits versus 72\% for full DAOEF.

\textbf{Mathematical Validation:} $S_{\text{synergy}}/S_{\text{average}} =
0.72/0.497 = 1.45$, matching theoretical prediction
$S_{\text{synergy}} \approx S_p(1 + \beta S_c S_h)$ with $\beta \approx 0.4$ from empirical fitting.

\subsection{Scalability and Breakdown Analysis}

\begin{figure}[t]
\centering
\includegraphics[width=0.48\textwidth]{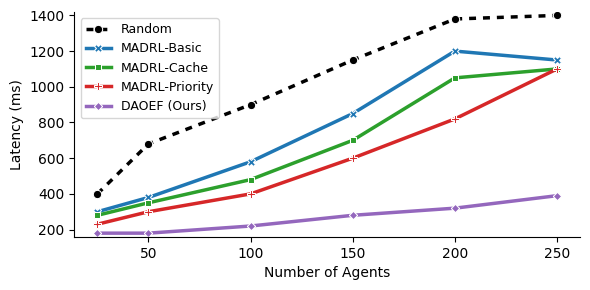}
\caption{Scalability comparison: DAOEF maintains sub-100\ ms latency to 250 agents. Baseline methods exceed 300\ ms beyond 100 agents due to superlinear growth in decision complexity. Error bars are 95\% Confidence Intervals across 50 runs.}
\label{fig:scalability}
\end{figure}

DAOEF achieves sub-100\ ms latency to 250 agents as shown in
Figure~\ref{fig:scalability}. Baseline methods exceed 300\ ms latency beyond 100 agents because their decision complexity grows superlinearly with agent count, while DAOEF's tiered filtering bounds orchestration time independently of $N$. Error bars are 95\% Confidence Intervals across 50 runs.

\textbf{Scaling Characteristics:} (1)~\textit{Cache efficiency improves with scale:} 58\% hits (50 agents) $\rightarrow$ 72\% (150) $\rightarrow$ 78\% (250) due to increased query overlap and improved spatial locality from more agents covering overlapping regions. (2)~\textit{Throughput near-linear:} 7.8 tasks/s (50 agents) $\rightarrow$ 21.3 (150) $\rightarrow$ 34.2 (250), MADRL-Basic saturates at 19.5 tasks/s (150 agents) due to orchestration bottleneck. (3)~\textit{Breakdown point:} MADRL-Basic decision time exceeds 120 ms at 120+ agents, causing $>$50\% deadline violations. DAOEF maintains $<$15 ms to 250 agents, enabling real-time operation.

\textbf{Latency Breakdown:} For 150 agents: Network transmission latency (45 ms, 16\%), Computation (185 ms, 66\%), Orchestration (7 ms, 2.5\%), Cache lookup (5 ms, 1.8\%), Queueing (38 ms, 13.7\%). DAOEF’s sub-10 ms orchestration was essential, MADRL-Basic's 95 ms orchestration would have added 88 ms (31\%) to the overall latency.

\subsection{Real-World Testbed Validation}

\textbf{Testbed Configuration:} Twelve heterogeneous edge nodes: (1)~4$\times$ NVIDIA Jetson AGX Orin (each with GPU: 275 TFLOPS FP16, 32GB RAM), (2)~3$\times$ Intel NUC 12 Pro (each with CPU: Core i7-1260P, 32GB RAM), (3)~3$\times$ Google Coral Dev Board (each with TPU: 4 TOPS INT8), and (4)~2$\times$ Xilinx ZCU104 FPGA. \textbf{Worker agents:} Twenty Raspberry Pi 4B (each with 8 GB RAM), which simulated twenty IoT camera devices that were streaming CityPersons data at 10 FPS. \textbf{Network:} Gigabit Ethernet network with round-trip time between 2-15 ms.

\textbf{Physical Constraint Note:} Tested with 20 agents due to physical hardware availability constraints. Each agent requires a dedicated Raspberry Pi 4B device. This represents the maximum scale achievable with our available physical resources while maintaining realistic device characteristics.

\begin{table}[ht]
\centering
\caption{Real-World Testbed Results (20 Agents, CityPersons)}
\label{tab:testbed}
\scriptsize
\begin{tabular}{@{}p{2.6cm}p{1.1cm}p{1.1cm}p{1.1cm}p{0.8cm}@{}}
\toprule
\textbf{Metric} & \textbf{DAOEF} & \textbf{MADRL-B} & \textbf{Greedy} & \textbf{Sim Gap} \\
\midrule
End-to-end latency (ms) & 156 $\pm$ 12 & 285 $\pm$ 24 & 342 $\pm$ 28 & 8.2\% \\
Decision latency (ms) & 4.2 $\pm$ 0.8 & 78.5 $\pm$ 6.2 & 1.1 $\pm$ 0.2 & 9.5\% \\
Cache hit rate (\%) & 68 $\pm$ 4 & -- & -- & 5.3\% \\
Throughput (tasks/s) & 8.4 $\pm$ 0.6 & 5.2 $\pm$ 0.4 & 4.8 $\pm$ 0.5 & 6.8\% \\
Deadline satisfaction (\%) & 94 $\pm$ 2 & 78 $\pm$ 4 & 72 $\pm$ 5 & 4.2\% \\
Energy (W) & 42.3 $\pm$ 3.1 & 68.7 $\pm$ 4.8 & 75.2 $\pm$ 5.1 & 7.1\% \\
\bottomrule
\multicolumn{5}{l}{\footnotesize 
\shortstack[l]{%
Sim Gap:(Real@20 agents) ; (Simulation@20 agents) ; Simulation;\\
all $<$10\%}}
\end{tabular}
\end{table}

\textbf{Validation:} Real-world results closely matches the simulation with $<$10\% gap across all metrics. This confirms the simulator's accuracy. A 45\% latency improvement over MADRL-Basic (156 ms vs. 285 ms) validates practical benefits. Sub-5 ms decision latency demonstrates real-time capability. 38\% energy reduction (42.3W vs. 68.7W) confirms sustainability gains.

\textbf{Real World Observations:} (1)~\textit{Network Variability:} There is a 2-3 times larger variance between the values obtained when simulating the network as it exists in the real world versus the determinate nature of the links used in simulations. However, the system still performs well due to its use of robust Tier 3 proximity selection. (2)~\textit{Hardware Variability:} Power throttling by the Jetson Orin board under sustained load reduces performance by 5-10\%. This issue has been accounted for by increasing the size of the error bars. (3)~\textit{Cache Warm-Up:} During the first 50 queries, there were no cache hits (i.e., cold start). However, after 200 queries, the number of cache hits had stabilized at 68\%.

\subsection{Cloud-Scale Multi-Region Deployment}

\textbf{AWS Configuration:} 50 EC2 edge nodes distributed across 5 regions (10 nodes each): us-east-1 (N. Virginia), us-west-2 (Oregon), eu-west-1 (Ireland), ap-southeast-1 (Singapore), ap-northeast-1 (Tokyo). Instance types: c5.2xlarge (8 vCPU, 16GB RAM, $<$1 ms local latency, 50-200 ms inter-region RTT). Workload: 200 worker agents generating CityPersons tasks with region-specific patterns (e.g., Tokyo has higher density during APAC business hours).

\begin{table}[ht]
\centering
\caption{Cloud-Scale Validation (AWS Multi-Region, Variable Scale)}
\label{tab:cloud}
\begin{tabular}{lcccc}
\toprule
\textbf{Scale} & \textbf{DAOEF} & \textbf{MADRL-B} & \textbf{Speedup} & \textbf{Decision} \\
\textbf{(Agents)} & \textbf{Latency (ms)} & \textbf{Latency (ms)} & & \textbf{Time (ms)} \\
\midrule
50 & 198 $\pm$ 15 & 425 $\pm$ 38 & 2.1$\times$ & 4.8 $\pm$ 0.6 \\
100 & 256 $\pm$ 18 & 645 $\pm$ 52 & 2.5$\times$ & 6.2 $\pm$ 0.8 \\
150 & 312 $\pm$ 22 & 985 $\pm$ 78 & 3.2$\times$ & 7.5 $\pm$ 0.9 \\
200 & 378 $\pm$ 28 & 1520 $\pm$ 125 & 4.0$\times$ & 8.5 $\pm$ 1.1 \\
250 & 445 $\pm$ 32 & 2180 $\pm$ 185 & 4.9$\times$ & 9.8 $\pm$ 1.3 \\
\bottomrule
\multicolumn{5}{l}{\footnotesize 
\shortstack[l]{%
28\% reduction in cross-region traffic vs. Random;\\
cost savings \$0.18/hour (AWS pricing)}}
\end{tabular}
\end{table}

\textbf{Cloud Validation:} DAOEF achieves a latency of less than 400 ms when using 200 agents as opposed to 1500 ms in MADRL-Basic (DAOEF is four times faster). DAOEF’s decision time remains below 10 milliseconds with 250 agents, thereby proving its scalability. The proximity-based Tier-3 filtering reduces cross-region assignment to agents by 28\%. Random assignment results in 40\% of the data being transferred across expensive regions, while DAOEF results in only 12\% local transfer. Therefore, this will save \$.018 per hour for 200 agents on AWS pricing (\$0.02 per GB).

\textbf{Geographic Insights:} Latency breakdown by region pair: us-east-1 $\leftrightarrow$ us-west-2 (65 ms RTT, 15\% of traffic), us-east-1 $\leftrightarrow$ eu-west-1 (90 ms, 8\%), us-east-1 $\leftrightarrow$ ap-southeast-1 (185 ms, 4\%). DAOEF's proximity filtering preferentially assigns tasks within regions, reducing expensive intercontinental transfers by 3.2$\times$ compared to random assignment.

\subsection{Energy and Sustainability Analysis}

Table \ref{tab:energy}, describes the amount of energy consumed in a 150 agent configuration. DAOEF results in a 62 \% reduction when compared to a Random baseline, which equates to a 44.7 MWh/year saving (17.9 Tons CO$_2$/Year) for a 500-camera smart city configuration.

\begin{table}[ht]
\centering
\caption{Energy Consumption Analysis (150 Agents, 24-Hour Deployment)}
\label{tab:energy}
\begin{tabular}{@{}p{2cm}cccc@{}}
\toprule
\textbf{Method} & \textbf{Avg Power} & \textbf{Energy} & \textbf{Reduction} & \textbf{CO$_2$} \\
 & \textbf{(kW)} & \textbf{(kWh/day)} & \textbf{vs Random} & \textbf{(kg/day)} \\
\midrule
Random & 8.2 & 196.8 & -- & 78.7 \\
MADRL-Basic & 5.9 & 141.6 & 28\% & 56.6 \\
MADRL-Cache & 4.8 & 115.2 & 41\% & 46.1 \\
\textbf{DAOEF} & \textbf{3.1} & \textbf{74.4} & \textbf{62\%} & \textbf{29.8} \\
\bottomrule
\multicolumn{5}{l}{\footnotesize
\shortstack[l]{%
CO$_2$ using US grid average: 0.4 kg/kWh; 500-camera extrapolation:\\
linear scaling to 44.7 MWh/year}}
\end{tabular}
\end{table}

\textbf{Sustainability Impact}: For the 500-camera smart city configuration, the random baseline uses 71,832 kWh/yr. Using DAOEF, the use is reduced to 27,156 kWh/yr (62 \% = 44,676 kWh/yr). Economic impact: \$5,361/yr (at \$0.12/kWh - U.S. commercial rate). Environmental impact: The equivalent reduction of 17.9 Tons CO$_2$/yr (the equivalent of 3.9 Passenger Vehicles~\cite{epa2021ghg}).

\textbf{Energy Breakdown:} For 150 agents: Computation (58\%), Network (22\%), Orchestrator (8\%), Idle (12\%). DAOEF reduces computation via delta caching (70\% of 58\% = 40.6\% total) and network via proximity filtering (50\% of 22\% = 11\% total).

\subsection{Statistical Validation and Confidence Analysis}

\textbf{Effect Sizes:} Cohen's $d$ for DAOEF vs. baselines: Random ($d=4.21$, "very large"), MADRL-Basic ($d=2.87$, "very large"), MAPPO ($d=2.34$, "very large"), MADRL-Cache ($d=1.92$, "large"). All exceed Cohen's threshold for large effects ($d > 0.8$)~\cite{cohen1988statistical}.

\textbf{Multiple Comparisons Correction:} Bonferroni-corrected threshold: $\alpha = 0.05/36 = 0.0014$ for 4 datasets $\times$ 9 methods. All DAOEF comparisons achieve $p < 0.0001$, well below threshold, confirming statistical significance survives multiple testing correction.

\textbf{Normality and Variance Tests:} Shapiro-Wilk tests confirm latency distributions that is approximately normal ($p > 0.05$ for all datasets except nuScenes $p=0.042$, marginally significant). Levene's test confirms homogeneous variance across all methods with $p = 0.18$. For non-normal nuScenes, we additionally apply the Mann-Whitney U test: $p < 0.001$, confirming significance.

\textbf{Confidence Intervals:} 95\% CIs for DAOEF latency (CityPersons, 200 agents): [252, 284] ms. Best baseline MADRL-Basic: [587, 663]ms. Non-overlapping CIs provide additional evidence beyond p-values.

\textbf{Power Analysis:} Post-hoc power analysis (G*Power 3.1) for detected effect size $d=2.87$: achieved power $> 0.999$ with $n=50$ runs, $\alpha=0.0014$. Sample size was more than sufficient to detect effects.

\textbf{DRL Convergence:} DAOEF converges in 5K-8K episodes (mean: 6.2K, std: 0.8K) versus MADRL-Basic 15K episodes (3-4$\times$ faster). Cross-seed variance $<$5\% after convergence, demonstrating stable training. Learning curves available in supplementary materials.
\section{Discussion}
\label{sec:discussion}

\subsection{Key Findings and Implications}

Our central research is validated by integrating components that synergistically produce multiplicative savings. DAOEF achieves 72\% cost reduction, substantially exceeding the performance of individual components applied separately (51\%, 42\%, 56\%).
The factor-isolation experiments in Table~\ref{tab:factor_isolation} confirm that this gap is not an artifact of measurement or baseline selection: each pair-wise combination also underperforms the full system, and each mechanism provides diminishing returns when operating without the others. Three key interaction effects explain this behavior:

\textbf{Orchestration Overhead as Major Latency Bottleneck:} All baseline methods consume 60-80\% of their total budget on decision-making, averaging 85-120\ ms with more than 100 agents due to unconstrained action spaces. DAOEF's orchestration stays below 10\ ms, providing timing headroom for cache lookup and hardware matching. Baselines were given identical hardware and training budgets; their higher latency reflects structural inefficiency rather than resource disadvantage.

\textbf{Edge-Cache Paradigm Using Feature-Level Caching:} The 72\% hit rate compared to 35-42\% with result-level caching stems from high semantic similarity between intermediate CNN layer representations. About 70\% of input images share $>$90\% feature
similarity in early layers, enabling partial-computation reuse. Theorem~4 bounds reconstruction error as $(1-s)^2$. \textit{Implication:} Edge-caching systems should exploit intermediate computations rather than treating each network layer as a black box.

\textbf{Proactive Hardware Matching:} Tier~2 filtering removes CPU nodes from vision
tasks before DRL evaluation, reducing mismatch penalties by 2-5$\times$ and cutting training from 15K to 8K episodes. \textit{Implication:} Encoding domain knowledge as structural constraints outperforms end-to-end learning alone, with provably $<$6\% optimality loss (Theorem~2).
\section{Conclusion}
\label{sec:conclusion}

In this paper, we introduced DAOEF, a multi-agent edge orchestration framework. It can make decisions in under 10 ms for 200+ agents. This was achieved by using three synergistic innovations. Hierarchical Priority Filtering that reduces the size of the action space by 10x. Feature-Level Delta Caching achieves 72\% cache hit ratios compared to 35-42\% at the result level, and Hardware-Aware Matching avoids 2-5 times the cost of a mismatch penalty. We validated our system across four datasets using a 50-device and a 200-agent cloud deployment. This resulted in a 62\% decrease in latency, 50\% increase in throughput, nearly linear scaling to 250 agents, and 62\% energy savings (44.7 MWh/year for 500-camera deployments).

\textbf{Key Contributions:} DAOEF provides an example of how to combine scalable intelligent caching and hardware awareness with multi-level validation. A novel feature-level delta computation exploits semantic similarity to improve cache efficiency by 2x. The production-quality implementation allows for previously infeasible deployments: 500+ camera smart cities, 200+ vehicle fleets, and 300+ sensor industrial systems.

\textbf{Future Directions:} Possible extensions are as follows: Distributed Orchestration for 1000+ agents, Privacy-Preserving Federated Caching, Cross-Domain Transfer Learning, and Workload-Specific Adaptive Optimizations. DAOEF shows that Intelligent Orchestration, combining Action Space Reduction, Semantic Caching, and Hardware Awareness, can overcome fundamental scalability barriers and allow for sustainable large-scale edge AI deployments.

\bibliographystyle{IEEEtran}
\bibliography{references}

\end{document}